\pdfoutput=1

\documentclass{article}
\usepackage{ijcai19}

\usepackage{times}
\usepackage{soul}
\usepackage{url}
\usepackage[utf8]{inputenc}
\usepackage[small]{caption}
\usepackage{graphicx}
\usepackage{amsmath}
\usepackage{booktabs}
\usepackage{algorithmic}
\usepackage[linesnumbered,ruled]{algorithm2e}
\usepackage{multirow}
\usepackage{caption}

\usepackage{aecompl}
\usepackage{amsfonts}
\usepackage{courier}  %Required
\usepackage[export]{adjustbox}
\usepackage{mathtools}
\usepackage{epstopdf}
\usepackage{float}
\usepackage{subfig}
\usepackage{xcolor}
\usepackage{diagbox}
\usepackage{graphbox}
\usepackage{multirow}
\usepackage{etoolbox}
\usepackage{caption}
\usepackage{kantlipsum}
\usepackage{hyperref}

\usepackage{float}
\usepackage{subfig}
\title{Adversarial Examples on Graph Data: Deep Insights into Attack and Defense}

\author{
Huijun Wu$^{1, 2}$
\and
Chen Wang$^2$\and
Yuriy Tyshetskiy $^{2}$\and
Andrew Docherty$^2$\and \\
Kai Lu$^3$ \and
Liming Zhu$^{1, 2}$\\
\affiliations
$^1$University of New South Wales, Australia\\
$^2$Data61, CSIRO\\
$^3$National University of Defense Technology, China
\emails
\{first, second\}@data61.csiro.au,
kailu@nudt.edu.cn
}

\begin{document}

\maketitle

\begin{abstract}
Graph deep learning models, such as graph convolutional networks (GCN) achieve remarkable performance for tasks on graph data. Similar to other types of deep models, graph deep learning models often suffer from adversarial attacks. However, compared with non-graph data, the discrete features, graph connections and different definitions of imperceptible perturbations bring unique challenges and opportunities for the adversarial attacks and defenses for graph data. In this paper, we propose both attack and defense techniques. For attack, we show that the discreteness problem could easily be resolved by introducing integrated gradients which could accurately reflect the effect of perturbing certain features or edges while still benefiting from the parallel computations. For defense, we observe that the adversarially manipulated graph for the targeted attack differs from normal graphs statistically. Based on this observation, we propose a defense approach which inspects the graph and recovers the potential adversarial perturbation. Our experiments on a number of datasets show the effectiveness of the proposed methods.
\end{abstract}

\section{Introduction}

Graph is commonly used to model many real-world relationships, such as social networks~\cite{newman2002random}, citation networks and transactions~\cite{ron2013quantitative} and the control-flow of programs~\cite{allen1970control}. The recent advance~\cite{kipf2016semi,velivckovic2017graph,cao2016deep,henaff2015deep} in deep learning expands its applications on graph data. One common task on graph data is \textit{node classification}: for a graph and labels of a portion of nodes, the goal is to predict the labels for the unlabelled nodes. This can be used to classify the unknown roles in the graph. For example, topics of papers in the citation network, customer types in the recommendation systems. 

Compared with the classic methods~\cite{bhagat2011node,xu2013node}, deep learning starts to push forward the performance of node classification tasks. The graph convolutional networks~\cite{bruna2013spectral,edwards2016graph} and its recent variants~\cite{kipf2016semi}  perform convolution operations in the graph domain by aggregating and combining the information of neighbor nodes. In these works, both node features and the graph structures (i.e., edges) are considered for classifying nodes. 

Deep learning methods are often criticized for their lack of robustness~\cite{goodfellow2014explaining}. In other words, it is not difficult to craft adversarial examples by only perturbing a tiny portion of examples to fool the deep neural networks to give incorrect predictions. Graph convolutional networks are no exception. These vulnerabilities under adversarial attacks are major obstacles for deep learning applications to be used in the safety-critical scenarios. In graph neural networks, one node can be a user in the social network or an e-commerce website. A malicious user may manipulate his profile or connect to targeted users on purpose to mislead the analytics system. Similarly, adding fake comments to specific products can fool the recommender systems of a website. 

The key challenge for simply adopting existing adversarial attack techniques used in non-graph data on graph convolutional networks is the discrete input problems. Specifically, the features of the graph nodes are often discrete. The edges, especially those in unweighted graphs, are also discrete. To address this, some recent studies have proposed greedy methods~\cite{wang2018attack,zugner2018adversarial} to attack the graph-based deep learning systems. A greedy method to perturb either features or graph structure iteratively. Graph structure and features statistics are preserved during the greedy attack. In this paper, we show that although having the discrete input issue, the gradients can still be approximated accurately by integrated gradients. Integrated gradients approximate Shapley values~\cite{hart1989shapley,lundberg2016unexpected} by integrating partial
gradients with respect to input features from reference input to the actual input. Integrated gradients greatly improve the efficiency of the node and edge selection in comparison to iterative methods. 

Compared with explorations in attacks, the defense of adversarial examples in graph models is not well-studied. In this paper, we show that one key reason for the vulnerabilities of graph models, such as GCN, is that these models are essentially aggregating the features according to graph structures. They heavily rely on the nearest neighboring information while making predictions on target nodes. We looked into the perturbations made by the existing attack techniques and found that adding edges which connect to nodes with different features plays the key role in all of the attack methods. In this paper, we show that simply performing pre-processing to the adjacency matrix of the graph is able to identify the manipulated edges. For nodes with bag-of-words (BOW) features, the Jaccard index is effective while measuring the similarities between connected nodes. By removing edges that connect very dissimilar nodes, we are able to defend the targeted adversarial attacks without decreasing the accuracy of the GCN models. Our results on a number of real-world datasets show the effectiveness and efficiency of the proposed attack and defense. 

%\textbf{We should emphasize that greedy or reinforcement learning etc. are expensive}

\section{Preliminaries}
\subsection{Graph Convolutional Network}
Given an attributed graph $\mathcal{G = (A, X)}$, $\mathcal{A} \in [0, 1]^{N \times N}$ is the adjacency matrix and $\mathcal{X} \in [0, 1]^{D}$ represents the $D$-dimenisonal binary node features. Assuming the indices for nodes and features are $\mathcal{V} = \{1, 2, ..., N\}$ and $\mathcal{F} = \{1, 2, ..., D\}$, respectively. We then consider the task of semi-supervised node classification where a subset of nodes $\mathcal{V_{L} \subseteq \mathcal{V}}$ are labelled with labels from classes $\mathcal{C} = \{1, 2,..., c_{K}\}$. The target of the task is to map each node in the graph to a class label. This is often called \textit{transductive} learning given the fact that the test nodes are already known during the training time.  

In this work, we study Graph Convolutional Network (GCN)~\cite{kipf2016semi}, a well-established method for semi-supervised node classifications. For GCN, initially, $H^{0} = X$. The GCN model then follows the following rule to aggregate the neighboring features:

\begin{equation}
    H^{(l + 1)}=\sigma(\tilde{D}^{-\frac{1}{2}}\tilde{A}\tilde{D}^{-\frac{1}{2}}H^{(l)}W^{(l)})
\end{equation}

where $\tilde{A} = A + I_{N}$ is the adjacency matrix of the graph $\mathcal{G}$ with self connections added, $\hat{D}$ is a diagonal matrix with $\tilde{D}_{i, i} = \Sigma_{j} \tilde{A}_{ij}$, and $\sigma$ is the activation function to introduce non-linearity. Each of the above equation corresponds to one graph convolution layer. A fully connected layer with softmax loss is usually used after $L$ layers of graph convolution layers for the classification. A two-layer GCN is commonly used for semi-supervised node classification tasks~\cite{kipf2016semi}. The model can, therefore, be described as:

\begin{equation}
    Z = f(X, A) = \text{softmax}(\hat{A}\sigma(\hat{A}XW^{(0)})W^{(1)})
\end{equation}

where $\hat{A} = \tilde{D}^{-\frac{1}{2}}\tilde{A}\tilde{D}^{-\frac{1}{2}}$. $\hat{A}$ is essentially the symmetrically normalized adjacency matrix. $W^{(0)}$ and $W^{(1)}$ are the input-to-hidden and hidden-to-output weights, respectively. 

\subsection{Gradients Based Adversarial Attacks}
Gradients are commonly exploited to attack deep learning models~\cite{yuan2019adversarial}. One can either use the gradients of the loss function or the gradients of the model output w.r.t the input data to achieve the attacks. Two examples are Fast Gradient Sign Method (FGSM) attack and Jacobian-based Saliency Map Approach (JSMA) attack. 
Fast Gradient Sign Method (FGSM)~\cite{ian2014adversarial} generates adversarial examples by performing gradient update along the direction of the sign of gradients of loss function w.r.t each pixel for image data. Their perturbation can be expressed as:

\begin{equation}
    \eta = \epsilon \text{sign}(\nabla J_{\theta}(x, l))
\end{equation}

where $\epsilon$ is the magnitude of the perturbation. The generated example is $x^{'} = x + \eta$. 

JSMA attack was first proposed in ~\cite{papernot2016limitations}. By exploiting the forward derivative of a DNN model, one can find the adversarial perturbations that force the model to misclassify the test point into a specific \textit{target class}. Given a feed-forward neural network $\mathbf{F}$ and sample $\mathbf{X}$, the Jacobian is computed by: 

\begin{equation}
    \nabla F(X) = \frac{\partial{F(X)}}{\partial{X}} = \left[ \frac{\partial{F_j(X)}}{\partial{x_{i}}} \right]_{i \in 1...M, j \in 1...N}    
\end{equation}

where the dimensions for the model output and input data are $M$ and $N$, respectively. To achieve a target class $t$, one wants $F_{t}(X)$ gets increased while $F_{j}(X)$ for all the other $j \neq t$ to decrease. This is accomplished by exploiting the adversarial saliency map which is defined by:

\begin{equation}
S(X, t)[i] = \left\{
\begin{array}{lr}
0,  \text{ if } \frac{\partial{F_t(X)}}{\partial{X_{i}}} < 0  \text{ or }  \Sigma_{j \neq t}{ \frac{\partial{F_j(X)}}{\partial{X_{i}}}} > 0 \\
\frac{\partial{F_t(X)}}{\partial{X_{i}}}|\Sigma_{j \neq t}{\frac{\partial{F_j(X)}}{\partial{X_{i}}}}| \text{, otherwise}\\
\end{array}
\right\}
\end{equation}

Starting from a normal example, the attacker follows the saliency map and iteratively perturb the example with a very tiny amount until the predicted label is flipped. For untargeted attack, one tries to minimize the prediction score for the winning class. 

\subsection{Defense for Adversarial Examples}
Although adversarial attack for a graph is a relatively new topic, a few works have been done as the defense for adversarial images on convolutional neural networks (e.g., ~\cite{xu2017feature,papernot2018deep}). For images, as the feature space is continuous, adversarial examples are carefully crafted with little perturbations. Therefore, in some cases, adding some randomization to the images is able to defend the attacks~\cite{xie2017mitigating}. Other forms of input pre-processing, such as local smoothing~\cite{xu2017feature} and image compression~\cite{shaham2018defending} have also been used to defend the attacks. These pre-processing works based on the observation that neighboring pixels of natural images are normally similar. Adversarial training~\cite{tramer2017ensemble} introduces the generated examples to the training data to enhance the robustness of the model. 

\section{Integrated Gradients Guided Attack}
Although FGSM and JSMA are not the most sophisticated attack techniques, they are still not well-studied for graph models. For image data, the success of FGSM and JSMA benefits from the continuous features in pixel color space. However, recent explorations in the graph adversarial attack techniques~\cite{zugner2018adversarial,dai2018adversarial} show that simply applying these methods may not lead to successful attacks. These work address this problem by either using greedy methods or reinforcement learning based methods which are often expensive.

The node features in a graph are often bag-of-words kind of features which can either be 1 or 0. The unweighted edges in a graph are also frequently used to express the existence of specific relationships, thus having only 1 or 0 in the adjacency matrix. When attacking the model, the adversarial perturbations are limited to either changing 1 to 0 or vice versa. The main issue of applying vanilla FGSM and JSMA in graph models is the inaccurate gradients. Given a target node $t$, for FGSM attack, $\nabla J_{W^{(1)}, W^{(2)}{}}(t) = \frac{\partial J_{W^{(1)}, W^{(2)}}(t)} {\partial{X}}$ measures the feature importance of all nodes to the loss function value. Here, $X$ is the feature matrix, each row of which describes the features for a node in the graph. For a specific feature $i$ of node $n$, a larger value of $\nabla {J_{W^{(1)}, W^{(2)}}}_{in}$ indicates perturbing feature $i$ to 1 is helpful to get the target node misclassified. However, following this gradient may not help for two reasons: First, the feature value might already be 1 so that we could not perturb it anymore; Second, even if the feature value is 0, since a GCN model may not learn a local linear function between 0 and 1 for this feature value, the result of this perturbation is unpredictable. It is also similar for JSMA as the Jacobian of the model shares all the limitations with the gradients of loss. In other words, vanilla gradients suffer from local gradient problems. Take a simple ReLU network $f(x) = ReLU(x)$ as an example, when $x$ increase from 0 to 1, the function value also increases by 1. However, computing the gradient at $x = 0$ gives 0, which does not capture the model behaviors accurately. 
To address this, we propose an integrated gradients based method rather than directly using vanilla derivatives for the attacks. Integrated gradients were initially proposed by~\cite{sundararajan2017axiomatic} to provide sensitivity and implementation invariance for feature attribution in the deep neural networks, particularly the convolutional neural networks for images.

The integrated gradient is defined as follows: for a given model $\mathbf{F}: R^{n} \rightarrow [0, 1]$, let $x \in R^{n}$ be the input, $x^{'}$ is the baseline input (e.g., the black image for image data). Consider a straight-line path from $x^{'}$ to the input $x$, the integrated gradients are obtained by accumulating all the gradients at all the points along the path. Formally, for the $i^{th}$ feature of $x$, the integrated gradients (IG) is as follows:

\begin{equation}
\small
IG_{i}(F(x)) ::= (x_{i} - x_{i}^{'}) \times \int_{\alpha = 0}^{1} \frac{\partial{F(x^{'} + \alpha x (x - x^{'}))}}{\partial{x_{i}}} d\alpha
\end{equation}

For GCN on graph data, we propose a generic attack framework. Given the adjacency matrix $A$, feature matrix $X$, and the target node $t$, we compute the integrated gradients for function $F_{W^{(1)}, W{(2)}}(A, X, t)$ w.r.t $I$ where $I$ is the input for attack. $I = A$ indicates edge attacks while $I = X$ indicates feature attacks. When $F$ is the loss function of the GCN model, we call this attack technique \emph{FGSM-like attack with integrated gradients}, namely IG-FGSM. Similarly, we call the attack technique by IG-JSMA when $F$ is the prediction output of the GCN model. For a targeted IG-JSMA or IG-FGSM attack, the optimization goal is to maximize the value of $F$. Therefore, for the features or edges having the value of 1, we select the features/edges which have the lowest negative IG scores and perturb them to 0. The untargeted IG-JSMA attack aims to minimize the prediction score for the winning class so that we try to increase the input dimensions with high IG scores to 0.

Note that unlike image feature attribution where the baseline input is the black image, we use the all-zero or all-one feature/adjacency matrices to represent the 1 $\rightarrow$ 0 or 0 $\rightarrow$ 1 perturbations. While removing a specific edge or setting a specific feature from 1 to 0, we set the adjacency matrix $\mathbf{A}$ and feature matrix $\mathbf{X}$ to all-zero respectively since we want to describe the overall change pattern of the target function $\mathbf{F}$ while gradually adding edges/features to the current state of $\mathbf{A}$ and $\mathbf{X}$. On the contrary, to add edges/features, we compute the change pattern by gradually removing edges/features from all-one to the current state, thus setting either $\mathbf{A}$ or $\mathbf{X}$ to an all-one matrix. To keep the direction of gradients consistent and ensure the computation is tractable, the IG (for edge attack) is computed as follows:

\begin{equation}
\scriptsize
IG(F(X, A, t))[i, j] \approx \left\{
\begin{array}{lr}
(A_{ij} - 0) \times \Sigma_{k = 1}^{m} \frac{\partial F(\frac{k}{m} \times (A_{ij} - 0))}{\partial{A_{ij}}} \times \frac{1}{m}, \\ \text{for removing edges}\\
(1- A_{ij}) \times \Sigma_{k = 1}^{m} \frac{\partial F(\frac{k}{m} \times (1 - A_{ij}))}{\partial{A_{ij}}} \times \frac{1}{m}, \\ \text{for adding edges}
\end{array}
\right.
\end{equation}

Algorithm~\ref{alg:ig_jsma} shows the pseudo-code for untargeted IG-JSMA attack. We compute the integrated gradients of the prediction score for winning class $c$ w.r.t the entries of $A$ and $X$. The integrated gradients are then used as metrics to measure the priority of perturbing specific features or edges in the graph $G$. Note that the edge and feature values are considered and only the scores of possible perturbations are computed (see Eq.(7)). For example, we only compute the importance of adding edges if the edge does not exist before. Therefore, for a feature or an edge with high perturbation priority, we perturb it by simply flipping it to a different binary value.

\begin{algorithm}[t]
\caption{IG-JSMA - Integrated Gradient Guided untargeted JSMA attack on GCN}
\label{alg:ig_jsma}
\footnotesize
 \SetKwInOut{Input}{Input}
 \Input{Graph $G^{(0)} = (A^{(0)}, X^{(0)})$, target node $v_{0}$ ~\\ 
 $F$: the GCN model trained on $G^{(0)}$ ~\\
 budget $\Delta$: the maximum number of perturbations.}
   \SetKwInOut{Output}{Output}
   \Output{Modified graph $G^{'} = (A^{'}, X^{'})$.}
 \SetKwFunction{algo}{Attack}
 \SetKwProg{myalg}{Procedure}{}{}
 \myalg{\algo{}}{
  //compute the gradients as the perturbation scores for edges and features. \quad\\
  $s_e$ $\leftarrow$ calculate$\_$edge$\_$importance(A) ~\\
  $s_f$ $\leftarrow$ calculate$\_$feature$\_$importance(X) ~\\
  //sort nodes and edges according to their scores. ~\\
  features $\leftarrow$ sort$\_$by$\_$importance(s$\_$f) ~\\
  edges  $\leftarrow$  sort$\_$by$\_$importance(s$\_$e) ~\\
  f $\leftarrow$ features.first, e $\leftarrow$ edges.first \\
  \While {$|A^{'} - A| + |X^{'} - X| < \Delta$}{
       //decide which to perturb ~\\
       \eIf {$s_e[e] > s_f[f]$} {
               flip feature f ~\\
               f $\leftarrow$ f.next} {
               flip edge e ~\\
               e = $\leftarrow$ e.next}
    }
    return $G_{'}$ 
    }
\end{algorithm}

While setting the number of steps $m$ for computing integrated gradients, one size does not fit all. Essentially, more steps are required to accurately estimate the discrete gradients when the function learned for certain features/edges is non-linear. Therefore, we enlarge the number of steps while attacking the nodes with low classification margins until stable performance is achieved. Moreover, the calculation can be done in an incremental way if we increase the number of steps by integer multiples.

To ensure the perturbations are unnoticeable, the graph structure and feature statistics should be preserved for edge attack and feature attack, respectively. The specific properties to preserve highly depend on the application requirements. For our IG based attacks,  we simply check against these application-level requirements while selecting an edge or a feature for perturbation. In practice, this process can be trivial as many statistics can be pre-computed or re-computed incrementally~\cite{zugner2018adversarial}.  

\section{Defense for Adversarial Graph}
In order to defend the adversarial targeted attacks on GCNs, we first hypothesize that the GCNs are easily attacked due to the fact that the GCN models strongly rely on the graph structure and local aggregations. The model trained on the attacked graph therefore suffers from the attack surface of the model crafted by the adversarial graph. As it is well known that adversarial attacks on deep learning systems are transferable to models with similar architecture and trained on the same dataset. Existing attacks on GCN models are successful as the attacked graphs are directly used to train the new model. Given that, one feasible defense is to make the adjacency matrix trainable. If the edge weights are learned during the training process, they may evolve so that the graph becomes different compared with the graph crafted by the adversary. 

We then verify this idea by making the edge weights trainable in GCN models. In CORA-ML dataset, we select a node that is correctly classified and has the highest prediction score for its ground-truth class. The adversarial graph was constructed by using nettack~\cite{zugner2018adversarial}. Without any defense, the target node is misclassified with the confidence of 0.998 after the attack. Our defense initializes the weights of the edges just as the adversarial graph. We then train the GCN model without making any additional modifications on the loss functions or other parameters of the model. Interestingly, with such a simple defense method, the target node is correctly classified with high confidence (0.912) after the attack. 

To explain why the defense works, we observe following the characteristics of the attacks: First, perturbing edges is more effective than modifying the features. This is consistent in all the attacks (i.e., FGSM, JSMA, nettack, and IG-JSMA). Feature-only perturbations generally fail to change the predicted class of the target node. Moreover, the attack approaches tend to favour adding edges over removing edges; Second, nodes with more neighbors are more difficult to attack than those with less neighbors. This is also consistent with the observations in ~\cite{zugner2018adversarial} that nodes with higher degrees have higher classification accuracy in both the clean and the attacked graphs.

Last, the attacks tend to connect the target node to nodes with different features and labels. We find out that this is the most powerful way to perform attacks. We verify this observation using CORA-ML dataset. To measure the similarity of the features, we use the Jaccard similarity score since the features of CORA-ML dataset are bag-of-words. Note that our defense mechanism is generic, while the similarity measures may vary among different datasets. For the graphs with other types of features, such as numeric features, we may use different similarity measures. Given two nodes $u$ and $v$ with $n$ binary features, the Jaccard similarity score measures the overlap that $u$ and $v$ share with their features. Each feature of $u$ and $v$ can either be 0 or 1. The total number of each combination of features for both $u$ and $v$ are specified as follows:

$M_{11}$ is the number of features where both $u$ and $v$ have a value of 1. $M_{01}$ is the feature number where the value of the feature is 0 in node $u$ but 1 in node $v$. Similarly, $M_{10}$ is the total number of features which have a value of 1 in node $u$ but 0 in node $v$. $M_{00}$ represents the total number of features which are 0 for both nodes. The Jaccard similarity score is given as

\begin{equation}
J_{u,v} = \frac{M_{11}}{M_{01} + M_{10} + M_{11}}.
\end{equation}

\begin{figure}[thb]
\centering
\subfloat[Clean]{
{\includegraphics[align=c, width=0.25\textwidth]{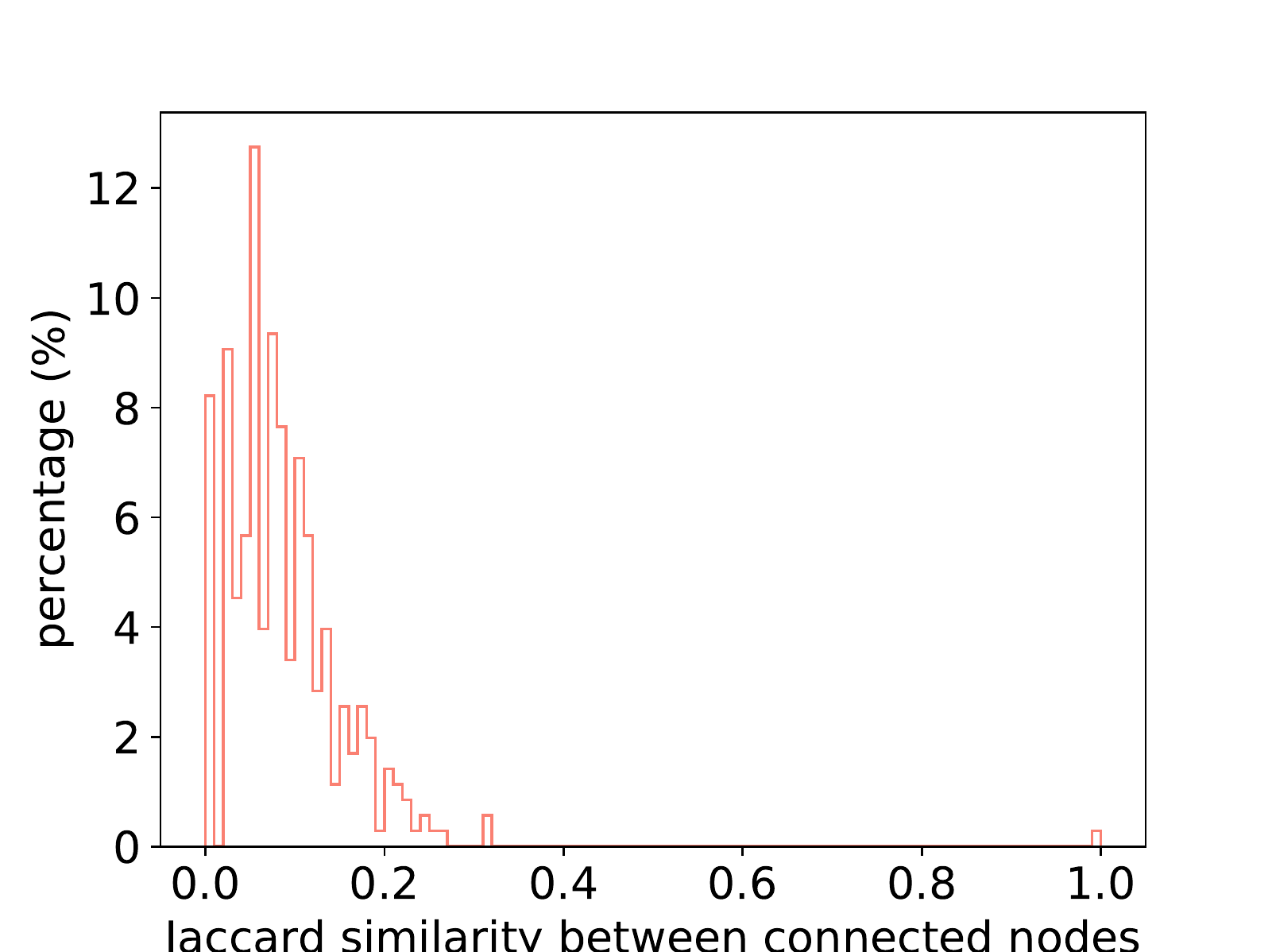}}
\label{fig:cora_margins}
}
\subfloat[Attacked]{{\includegraphics[align=c,width=0.25\textwidth]
{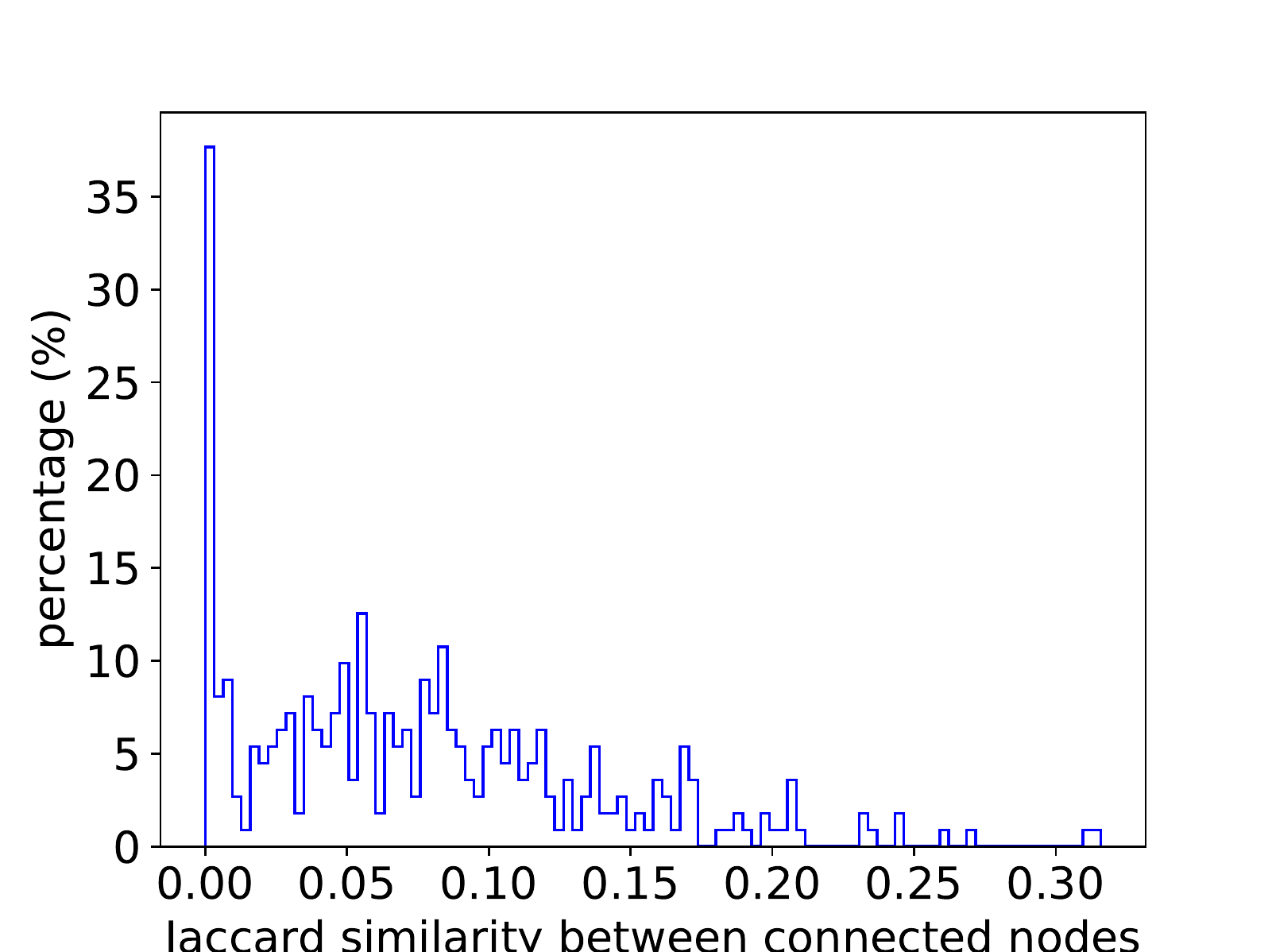}}
\label{fig:citeseer_margins}
}
\caption{Histograms for the Jaccard similarities between connected nodes before and after FGSM attack.}
\label{fig:jaccard_similarity}
\end{figure}

We train a two-layer GCN on the CORA-ML dataset and study the nodes that are classified correctly with high probability (i.e., $\geq$ 0.8). For these nodes, Figure~\ref{fig:jaccard_similarity} shows the histograms for the Jaccard similarity scores between connected nodes before and after the FGSM attack. The adversarial attack significantly increases the number of neighbors which have low similarity scores to the target nodes. This also stands for nettack~\cite{zugner2018adversarial}. For example, we enable both feature and edge attacks for nettack and attack the node 200 in the GCN model trained on CORA-ML dataset. Given the node degree of 3, the attack removes the edge 200 $\rightarrow$ 1582 because node 1582 and node 200 are similar ($J_{1582, 200}$ = 0.113). Meanwhile, the attacks add edge 200 $\rightarrow$ 1762 and 200 $\rightarrow$ 350, and node 200 shares no feature similarity with the two nodes. No features were perturbed in this experiment.

This result explains our observations. Compared with deep convolutional neural networks (for image data) which have often more layers and parameters than graph neural networks. GNNs, such as GCN for node classifications, are relatively simple. They essentially aggregate the features according to the graph structure. For a target node, an adversarially crafted graph attempts to connect the nodes with different features and labels to pollute the representation of the target node to make the target node less similar to the nodes within its correct class. Correspondingly, while removing edges, the attack tends to remove the edges connecting the nodes that share many similarities to the target node. The edge attacks are more effective due to the fact that adding or removing one edge  affects all the feature dimensions during the aggregation. In contrast, modifying one feature only affects one dimension in the feature vector and the perturbation can be easily masked by other neighbors of nodes with high degrees.

\begin{figure}[th]
\centering
\includegraphics[width=0.32\textwidth]{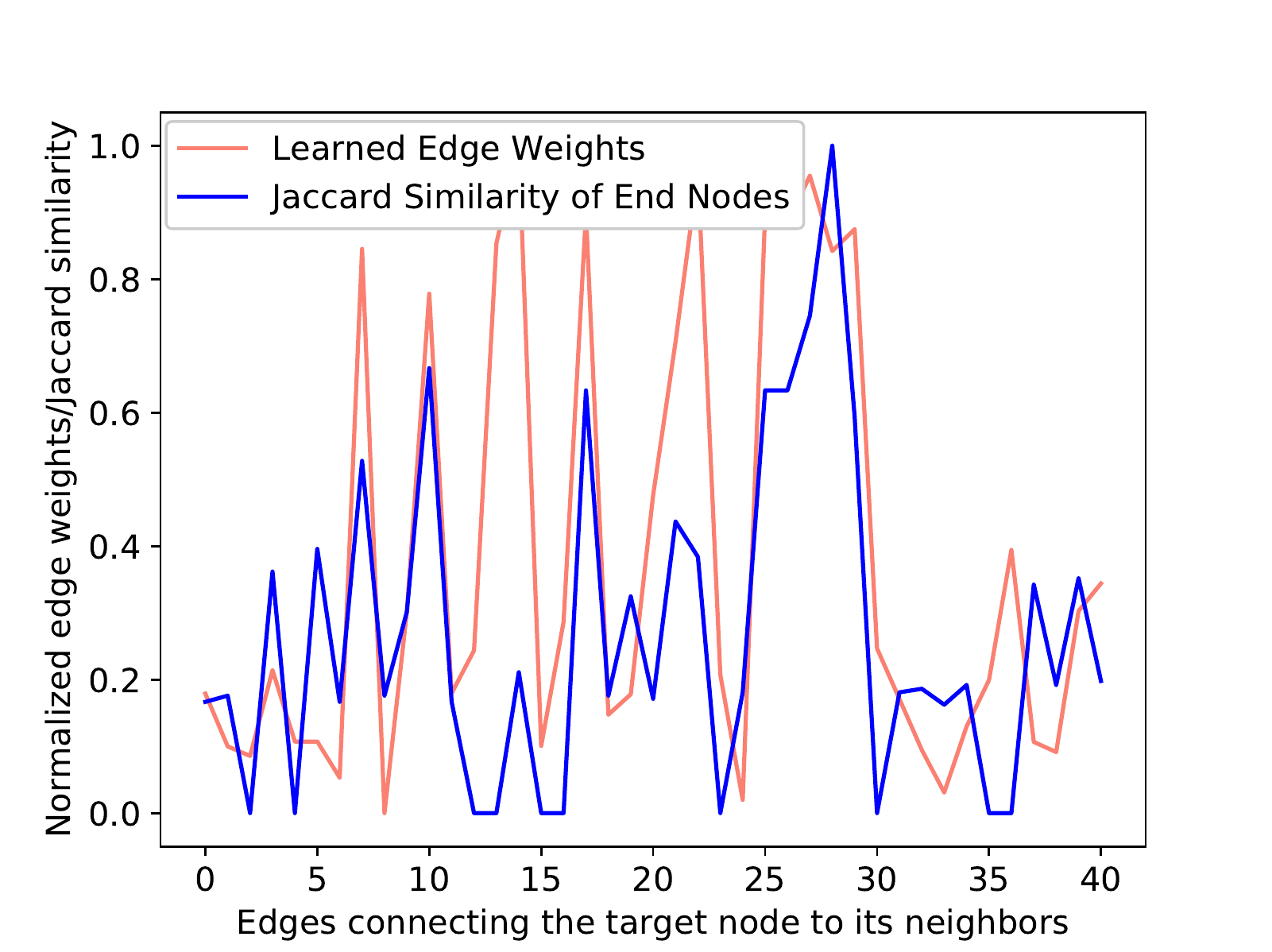}
\caption{The normalized learned edge weights and the Jaccard similarity scores for the end nodes of the edges. Each value of the x-axis represents an edge in the neighborhood of the target node.}
\label{fig:weights_vs_jaccard}
\end{figure}

Based on these observations, we make another hypothesis that the above defense approach works because the model assigns lower weights to the edges that connect the target node to the nodes sharing little feature similarity with it. To verify this, we plot the learned weights and the Jaccard similarity scores of the end nodes for the edges starting from the target node (see Figure~\ref{fig:weights_vs_jaccard}). Note that for the target node we choose, the Jaccard similarity scores between every neighbor of the target node and itself are larger than 0 in the clean graph. The edges with zero similarity scores are all added by the attack. As expected, the model learns low weights for most of the edges with low similarity scores.

To make the defense more efficient, we do not even need to use learnable edge weights as the defense. Learning the edge weights inevitably introduces extra parameters to the model, which may affect the its scalability and accuracy. A simple approach is potentially as effective based on the following: First, normal nodes generally do not connect to many nodes that share no similarities with it; Second, the learning process essentially assigns low weights to the edges connecting two dissimilar nodes. We therefore propose a simple yet effective defense approach based on the following insight. 

Our defense approach is pre-processing based. We perform a pre-processing on a given graph before training. We check the adjacency matrix of the graph and inspect the edges. All the edges that connect nodes with low similarity score (e.g., = 0) are selected as candidates to remove. Although the clean graph may also have a small number of such edges, we find that removing these edges does little harm to the prediction of the target node. On the contrary, the removal of these edges may improve the prediction in some cases. This is intuitive as aggregating features from nodes that differ sharply from the target often over-smooths the node representations. In fact, a recent study~\cite{sgc} shows that the nonlinearity and multiple weight matrices at different layers do not contribute much to the predictive capabilities of GCN models but introduce unnecessary complexity. \cite{zugner2018adversarial} uses a simplified surrogate model to achieve the attacks on GCN models for the same reason. Dai et al.~\cite{dai2018adversarial} briefly introduces a defense method by dropping some edges during the training. They show this decreases the attack rate slightly. In fact, their method works only when the edges connecting dissimilar nodes are removed. However, this defense fails to differentiate the useful edges from those need to be removed, thus achieving sub-optimal defense performance.

The proposed defense is computationally efficient as it only makes one pass to the existing edges in the graph, thus having the complexity of $O(N)$ where $N$ is the number of edges. For large graphs, calculating the similarity scores can be easily parallelized in implementation. 

\section{Evaluation}
We use the widely used CORA-ML, CITESEER~\cite{bojchevski2017deep} and Polblogs~\cite{adamic2005political} datasets. The overview of the datasets is listed below.

\begin{table}[H]
\centering
\scriptsize
\caption {Statistics of the datasets.} \label{tab:datasets} 
\begin{tabular}{@{}llll@{}}

\toprule

Dataset  & Nodes & Features & Edges \\ \midrule
CORA-ML  & 2708  & 1433     & 5429  \\
Citeseer & 3327  & 3703     & 4732  \\
Polblogs & 1490  & -        & 19025 \\ \bottomrule
\end{tabular}
\end{table}

We split each graph in labeled (20\%) and unlabeled nodes
(80\%). Among the labeled nodes, half of them is used for training while the rest half is used for validation. For the polblogs dataset, since there are no feature attributes, we set the attribute matrix to an identity matrix.

\subsection{Transductive Attack}
As mentioned, due to the transductive setting, the models are not regarded as fixed while attacking. After perturbing either features or edges, the model is retrained for evaluating the attack effectiveness. To verify the effectiveness of the attack, we select the nodes with different prediction scores. Specifically, we select in total 40 nodes which contain the 10 nodes with top scores, 10 nodes with the lowest scores and 20 randomly selected nodes. We compare the proposed IG-JSMA with several baselines including random attacks, FGSM, and nettack. Note that for the baselines, we conducted \textit{direct} attacks on the features of the target node or the edges directly connected to the target node. Direct attacks achieve much better attacks so that can act as stronger baselines. 

\begin{figure*}[thb]
\centering
\subfloat[CORA]{
{\includegraphics[align=c, width=0.28\textwidth]{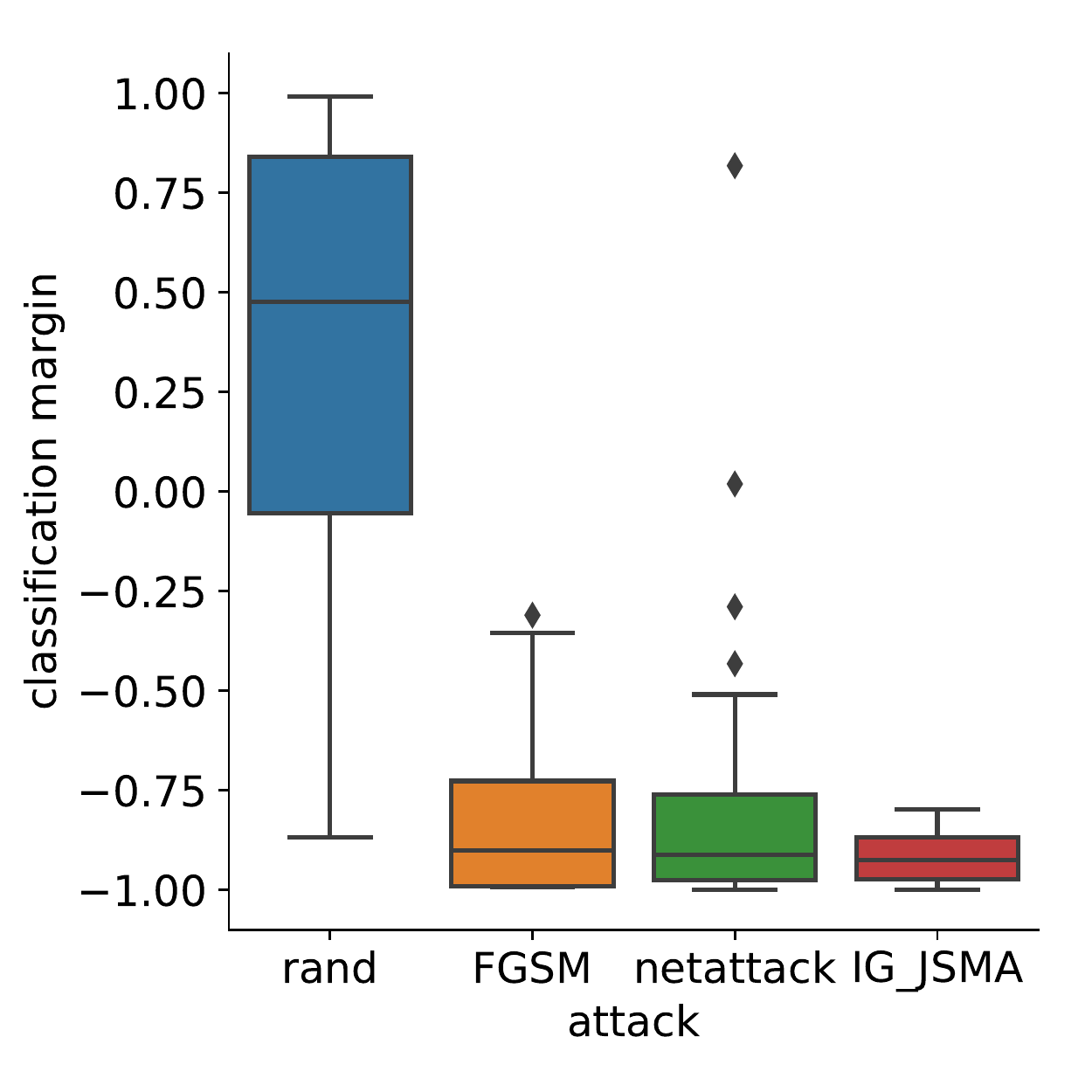}}
\label{fig:cora_margins}
}
\subfloat[Citeseer]{{\includegraphics[align=c,width=0.28\textwidth]
{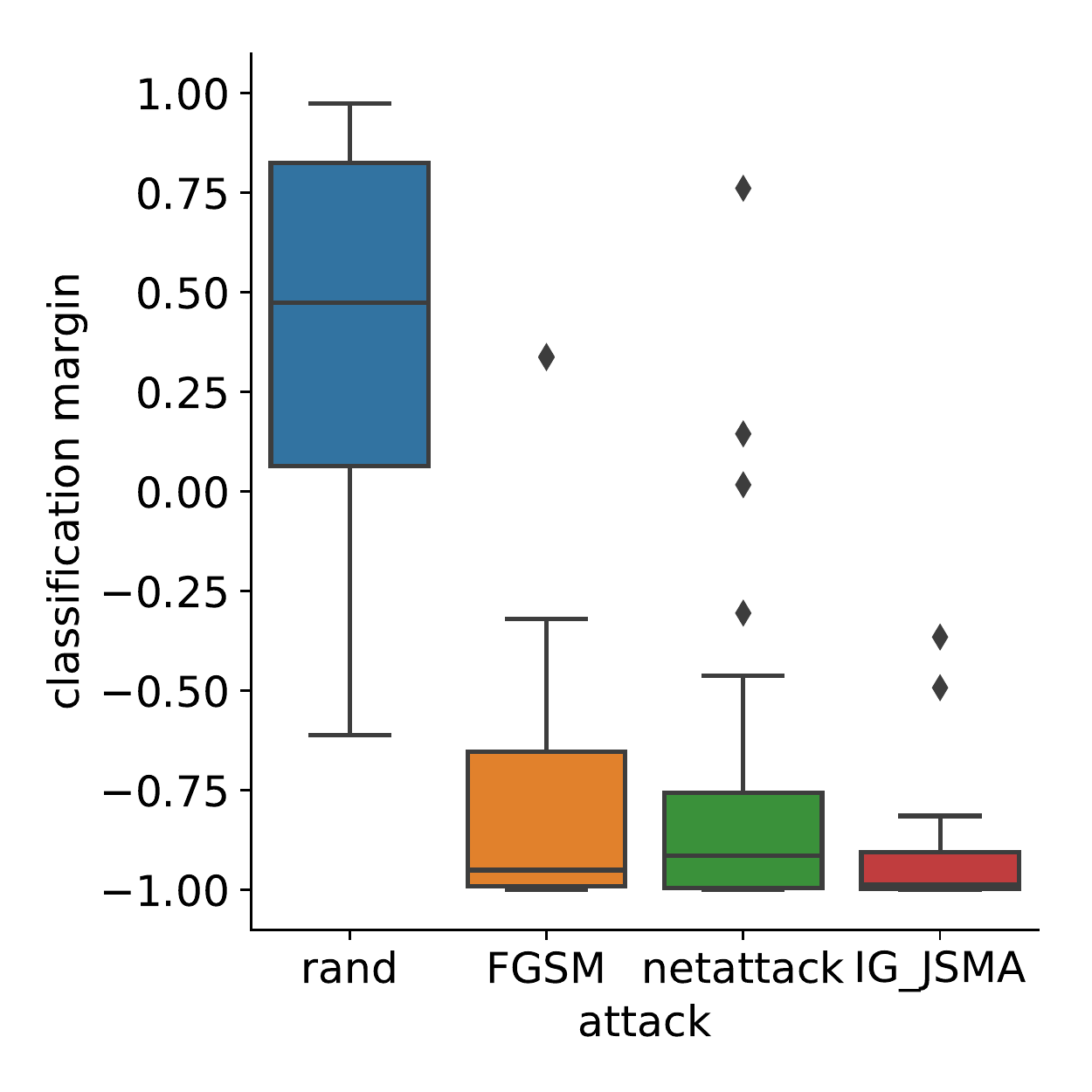}}
\label{fig:citeseer_margins}
}
\subfloat[polblogs]{{\includegraphics[align=c,width=0.28\textwidth]
{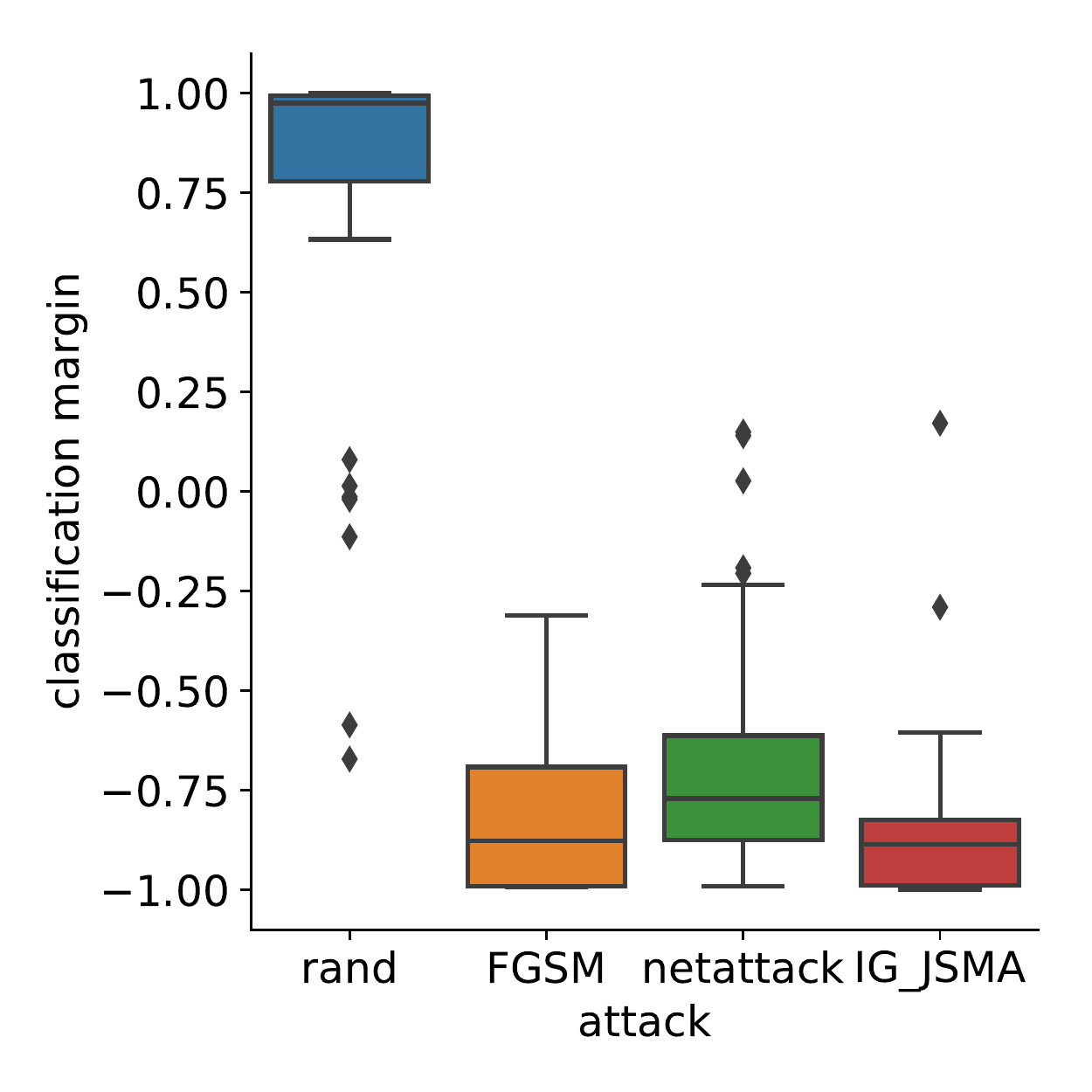}}
\label{fig:polblogs}
}

\caption{The classification margin under different attack techniques.}
\label{fig:classification_margin}
\end{figure*}

To evaluate how effective is the attack, we use classification margins as the metric. For a target node $v$, the classification margin of $v$ is $Z_{v, c} - max_{c^{'} \neq c}Z_{v, c^{'}}$ where $c$ is the ground truth class, $Z_{v, c}$ is the probability of class $c$ given to the node $v$ by the graph model. A lower classification margin indicates better attack performance. Figure~\ref{fig:classification_margin} shows the classification margins of nodes after re-training the model on the modified graph. We found that IG-JSMA outperforms the baselines. More remarkably, IG-JSMA is quite stable as the classification margins have much less variance. Just as stated in ~\cite{zugner2018adversarial}, the vanilla gradient-based methods, such as FGSM are not able to capture the actual change of loss for discrete data. Similarly, while used to describe the saliency map, the vanilla gradients are also not accurate.

To demonstrate the effectiveness of IG-JSMA, we also compare it with the original JSMA method where the saliency map is computed by the vanilla gradients. Table~\ref{tab:comparisons_jsma} compares the ratio of correctly classified nodes after the JSMA and IG-JSMA attacks for 100 random sampled nodes, respectively. A lower value is better as more nodes are misclassified. We can see that IG-JSMA outperforms JSMA attack. This shows that the saliency map given by integrated gradients approximate the change patterns of the discrete features/edges better.

\begin{table}[h]
\centering
\scriptsize
\caption {The ratio of correctly classified nodes under JSMA and IG-JSMA attacks.} \label{tab:comparisons_jsma} 
\begin{tabular}{@{}cccccc@{}}
\toprule
Dataset                & CORA & Citeseer & Polblogs \\ \midrule
JSMA            & 0.04   &  0.06     &  0.04  \\
IG\_JSMA & 0.00   & 0.01   &   0.01                   
 \\ \bottomrule
\end{tabular}
\end{table}

\begin{figure*}[h!]
\centering
\subfloat[Ground Truth]{
{\includegraphics[align=c, width=0.3\textwidth]{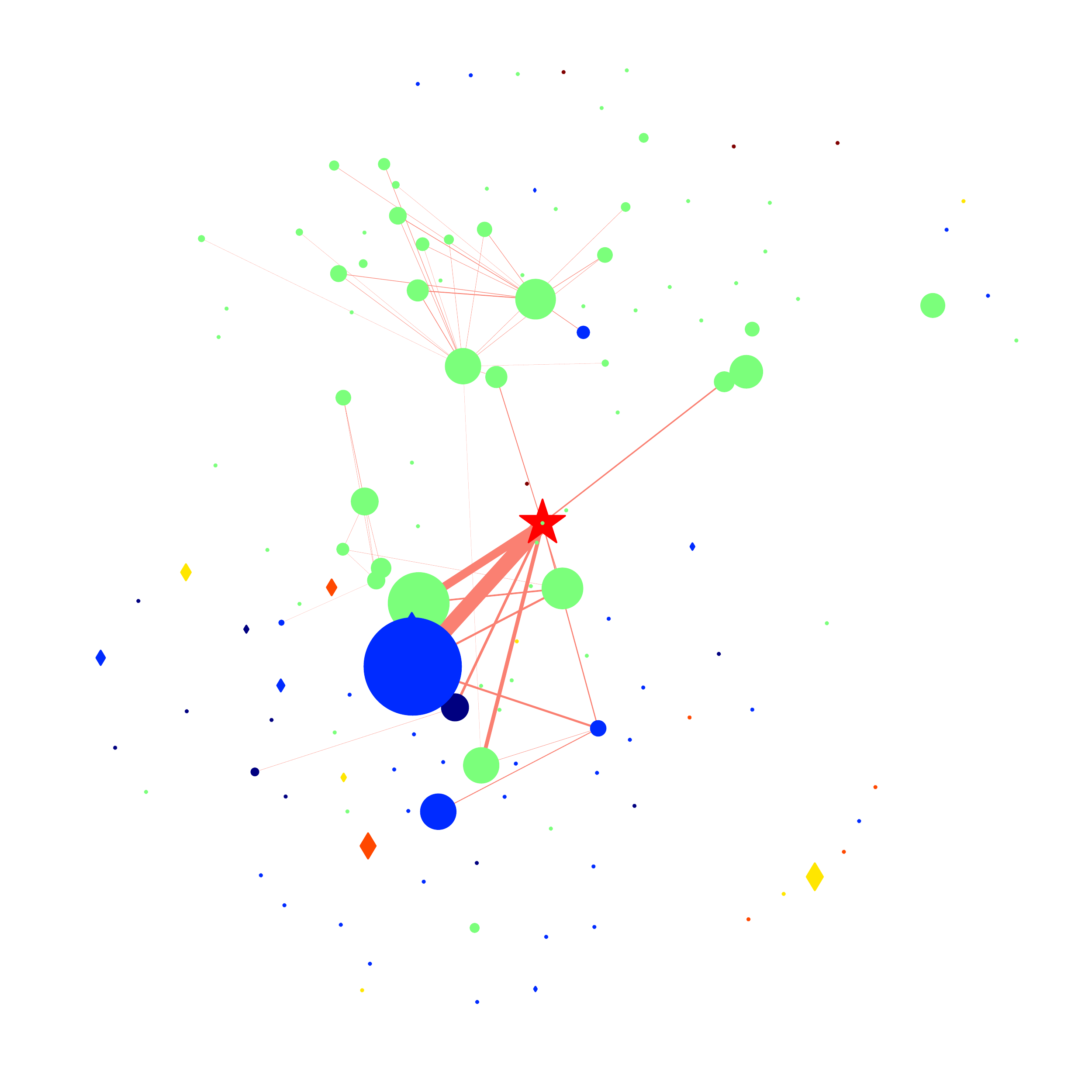}}
\label{fig:ground_truth}
}
\subfloat[Vanilla Gradients]{{\includegraphics[align=c,width=0.3\textwidth]
{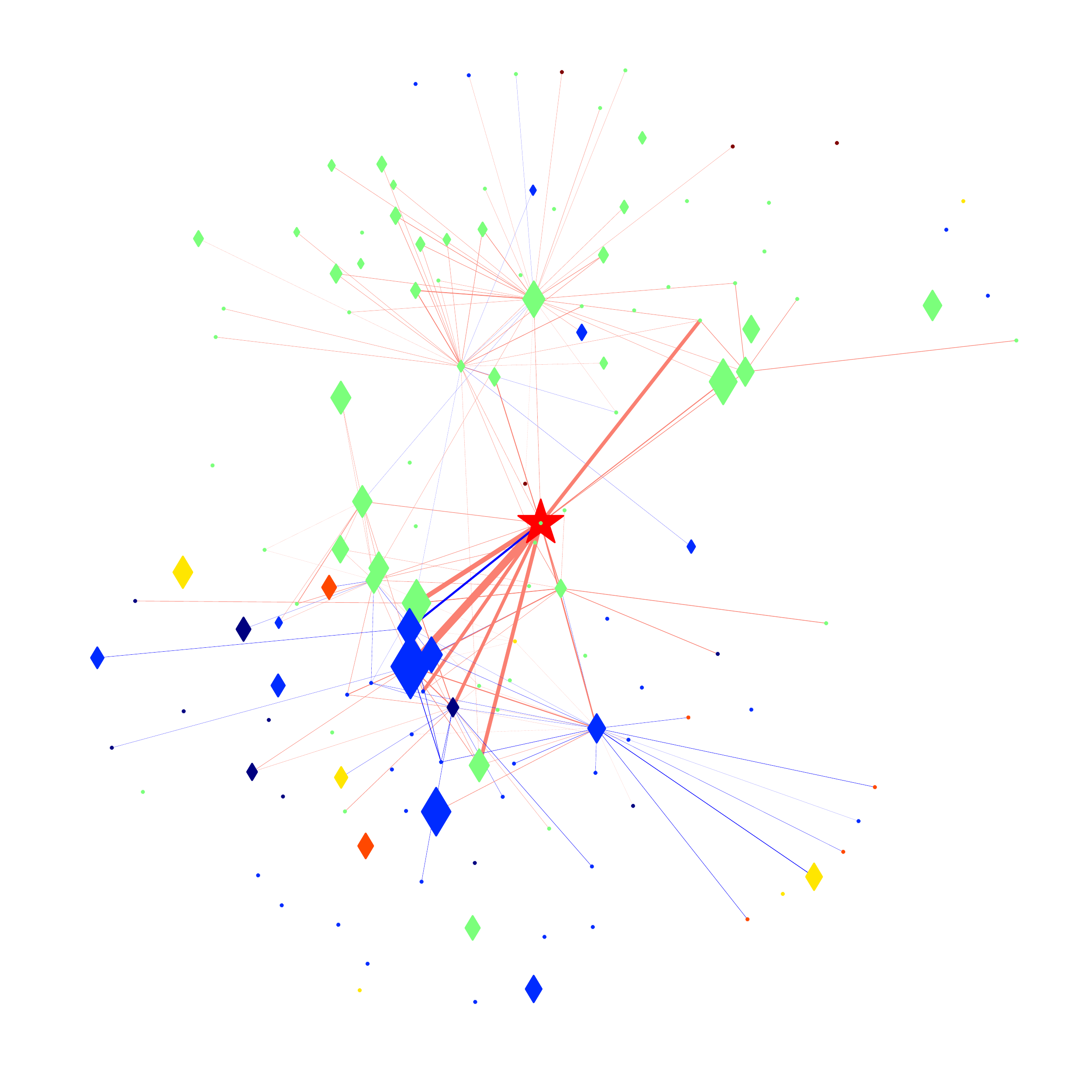}}
\label{fig:vanilla}
}
\subfloat[Integrated Gradients]{{\includegraphics[align=c,width=0.3\textwidth]
{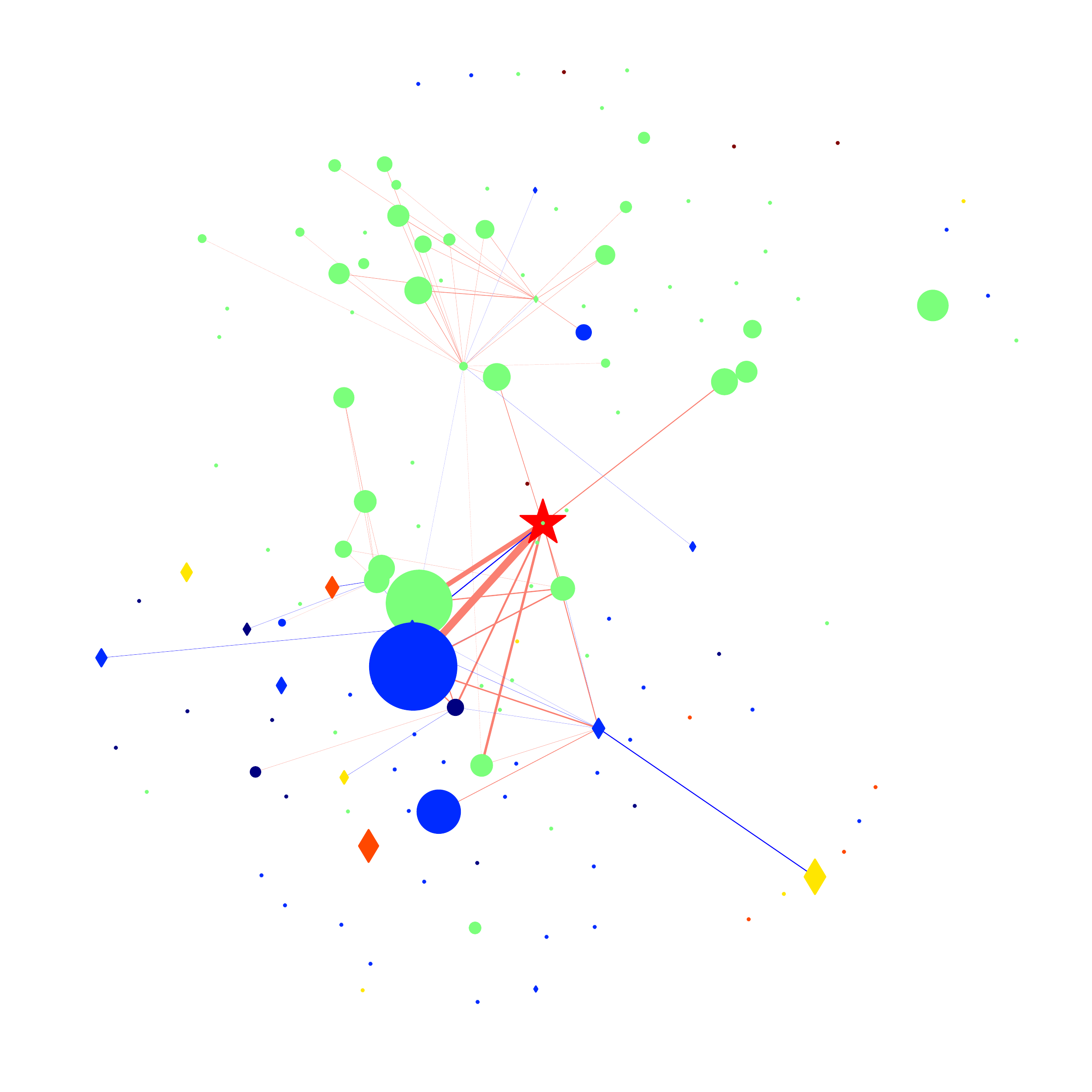}}
\label{fig:integrated}
}
\caption{The approximations of node/edge importance.}
\label{fig:node_link_importance}
\end{figure*}

Figure~\ref{fig:node_link_importance} gives an intuitive example about this. For the graph, we conducted evasion attack where the parameters of the model are kept fixed as the clean graph. For a target node in the graph, given a two-layer GCN model, the prediction of the target node only relies on its two-hop ego graph. We define the importance of a feature/an edge as follows: For a target node $v$, The brute-force method to measure the importance of the nodes and edges is to remove one node or one edge at a time in the graph and check the change of prediction score of the target node. 

Assume the prediction score for the winning class $c$ is $p_{c}$. After setting entry $\mathcal{A}_{ij}$ of the adjacency matrix from 1 to 0, the $p_{c}$ changes to $p_{c}^{'}$. We define the importance of the edge by $\Delta_{p_{c}} = p_{c}^{'} - p_{c}$. To measure the importance of a node, we could simply remove all the edges connected to the node and see how the prediction scores change. The importance values can be regarded as the ground truth discrete gradients. 

Both vanilla gradients and integrated gradients are approximations of the ground truth importance scores. The node importance can be approximated by the sum of the gradients of the prediction score w.r.t all the features of the node as well as the gradients w.r.t to the entries of the adjacency matrix.

In Figure~\ref{fig:node_link_importance}, the node color represents the class of the node. Round nodes indicate positive importance scores while diamond nodes indicate negative importance score. The node size indicates the value of the positive/negative importance score. A larger node means higher importance. Similarly, red edges are the edges which have positive importance scores while blue ones have negative importance scores. Thicker edges correspond to more important edges in the graph and the pentagram represents the target node in the attack.

Figure~\ref{fig:ground_truth}, ~\ref{fig:vanilla} and ~\ref{fig:integrated} show the node importance results of brute-force, vanilla gradients and integrated gradients approach respectively (\# of steps = 20). The vanilla gradients reveal little information about node/edge importance as almost all the edges are assigned with certain importance scores and it is difficult to see the actual node/edge influence. However, in the brute-force case, we notice that the majority number of edges are considered not important for the target node. Moreover, vanilla gradients underestimate the importance of nodes overall. The integrated gradients, as shown in Figure~\ref{fig:integrated} is consistent with the ground truth produced by brute-force approach shown in Figure~\ref{fig:ground_truth}. With only 20 steps along the path, integrated gradients provide accurate approximations for the importance scores. This shows the integrated gradients approach is effective when used to guide the adversarial attacks on graphs with discrete values.

\subsection{Defense}

In the following, we study the effectiveness of the proposed defense technique under different settings. We use the CORA-ML and Citeseer datasets that have features for the nodes. We first evaluate whether the proposed defense method affects the performance of the model. Table~\ref{tab:accu_defense} shows the accuracy of the GCN models with/without the defense. 

\begin{table}[h!]
\centering
\scriptsize
\caption {Accuracy (\%) of models on clean data with/without the proposed defense. We remove the outliers (i.e., accu $\leq 75\%/65\%$ for CORA-ML/Citeseer) due to the high variance.} \label{tab:accu_defense} 
\begin{tabular}{@{}lll@{}}
\toprule
Dataset  & w/o defensde & w/ defense \\ \midrule
CORA-ML  &     80.9$\pm$ 0.6         &     80.7 $\pm$ 0.7       \\
Citeseer &     69.5 $\pm$ 0.7         &     69.6 $\pm$ 0.8       \\ \bottomrule
\end{tabular}
\end{table}

\begin{table}[t]
\centering
\scriptsize
\caption {Classification margins and error rates (\%) for the GCN models with different attacks.} \label{tab:accu_defense} 

\begin{tabular}{@{}llllll@{}}
\toprule
\multirow{2}{*}{Dataset}  & \multirow{2}{*}{Attack} & \multicolumn{2}{l}{CM (w/ attack)} & \multicolumn{2}{l}{Accu (w/ attack)} \\ \cmidrule(l){3-6} 
                          &                         & w/ defense      &no defense      & w/ defense          &no defense          \\ \cmidrule(r){1-6}
\multirow{4}{*}{CORA}     & FGSM                    &  0.299 $\pm$ 0.741  &     -0.833 $\pm$ 0.210     &    0.625 &    0.025                   \\
                          & JSMA                    &  0.419 $\pm$ 0.567  &   -0.828 $\pm$ 0.225 &     0.775      &       0      \\
                          & nettack                 &  0.242 $\pm$ 0.728  &   -0.839 $\pm$ 0.343 &  0.600       &  0.025            \\
                          & IG-JSMA                 &   0.397 $\pm$ 0.553 &  -0.897 $\pm$  0.114  & 0.750    &  0               \\ \cmidrule(r){1-6}
\multirow{4}{*}{Citeseer} & FGSM                    &   0.451 $\pm$ 0.489 & -0.777 $\pm$ 0.279    & 0.825        & 0.025                      \\
                          & JSMA                    &    0.501 $\pm$ 0.531 &  -0.806 $\pm$ 0.186 & 0.775        &  0.05                     \\
                          & nettack                 &   0.421 $\pm$ 0.468  & -0.787 $\pm$ 0.332 &  0.775    &  0.025                    \\
                          & IG-JSMA                 &     0.495 $\pm$ 0.507  & -0.876 $\pm$ 0.186    & 0.800  & 0.025                      \\ \bottomrule
\end{tabular}
\end{table}

We find that the proposed defense was cheap to use as the pre-processing of our defense method almost makes no negative impact on the performance of the GCN models. Moreover, the time overhead is negligible. Enabling defense on the GCNs models for the two datasets increases the run time of training by only 7.52s and 3.79s, respectively. Note that run time results are obtained using our non-optimized Python implementation.

For different attacks, we then evaluate how the classification margins and accuracy of the attacked nodes change with/without the defense. As in the experiments of transductive attack, we select 40 nodes with different prediction scores. The statistics of the selected nodes are the followings: For CORA-ML and Citeseers datasets, we train the GCN models on the clean graphs. The selected nodes have classification margins of $0.693 \pm 0.340$ and $0.636 \pm 0.419$, respectively.

The results are given in Table~\ref{tab:accu_defense}. First of all, without defenses, most of the selected nodes are misclassified as the accuracy is always under 0.05 for any attacks. By enabling the defense approach, the accuracy can be significantly improved regardless of the attack methods. This, to some degree, shows that all the attack methods seek similar edges to attack and the proposed defense approach is attack-independent. Although a few nodes were still misclassified with the defense, the prediction confidence for their winning class is much lower since the classification margins increase. Therefore, it becomes harder to fool the users because manual checks are generally involved in predictions with low confidence. Overall, the proposed defense is effective even though we only remove the edges that connect nodes with Jaccard similarity score of 0.

\section{Conclusions and Discussion}
Graph neural networks (GNN) significantly improved the analytic performance on many types of graph data. However, like deep neural networks in other types of data, GNN suffers from robustness problems. In this paper, we gave insight into the robustness problem in graph convolutional networks (GCN). We proposed an integrated gradients based attack method that outperformed existing iterative and gradient-based techniques in terms of attack performance. We also analyzed attacks on GCN and revealed the robustness issue was rooted in the local aggregation in GCN. We give an effective defense method to improve the robustness of GCN models. We demonstrated the effectiveness and efficiency of our methods on benchmark data. %Although we use the GCN model as a case in this paper, both the attack and defense principles are applicable to other variations of GNNs due to the fact that these models are also aggregation-based.

\bibliographystyle{named}
\bibliography{ijcai19}

\end{document}